\documentclass[10pt,twocolumn,letterpaper]{article}

\usepackage{cvpr}
\usepackage{times}
\usepackage{epsfig}
\usepackage{graphicx}
\usepackage{inconsolata}

\usepackage{array}
\usepackage{booktabs}
\usepackage{caption}
\usepackage{subcaption}

 \makeatletter
 \DeclareRobustCommand\onedot{\futurelet\@let@token\@onedot}
 \def\@onedot{\ifx\@let@token.\else.\null\fi\xspace}
 \def\eg{e.g\onedot}

 \makeatother

\usepackage[pagebackref=true,breaklinks=true,letterpaper=true,colorlinks,bookmarks=false]{hyperref}

\cvprfinalcopy

\def\cvprPaperID{1114} %

\ifcvprfinal\fi
\begin{document}

\newcommand{\nmn}{Neural Module Networks\xspace}
\newcommand{\mod}[1]{{\small\texttt{#1}}}
\newcommand{\todo}{\textcolor{red}}
\newcommand{\shapes}{{\textsc{shapes}}\xspace}
\newcommand{\cocoqa}{{\textsc{CocoQA}}\xspace}
\newcommand{\marcus}[1]{\textcolor{blue}{Marcus: #1}}
\title{Deep Compositional Question Answering with Neural Module Networks}

\author{Jacob Andreas \hspace{1em} Marcus Rohrbach \hspace{1em} Trevor Darrell \hspace{1em} Dan Klein \\
Department of Electrical Engineering and Computer Sciences \\
University of California, Berkeley\\
{\tt\small \{jda,rohrbach,trevor,klein\}@\{cs,eecs,eecs,cs\}.berkeley.edu}
}

{
\renewcommand{\ttdefault}{pcr}
\maketitle
}
\thispagestyle{empty}
\begin{abstract}
  Visual question answering is fundamentally compositional in nature---a
  question like \emph{where is the dog?} shares substructure with questions like
  \emph{what color is the dog?} and \emph{where is the cat?} This paper seeks to
  simultaneously exploit the representational capacity of deep networks and the
  compositional linguistic structure of questions.  We describe a procedure for
  constructing and learning \emph{neural module networks}, which compose
  collections of jointly-trained neural ``modules'' into deep networks for
  question answering. Our approach decomposes questions into their linguistic
  substructures, and uses these structures to dynamically instantiate modular
  networks (with reusable components for recognizing dogs, classifying colors,
  etc.). The resulting compound networks are jointly trained. We evaluate our
  approach on two challenging datasets for visual question answering,
  achieving state-of-the-art results on both the VQA natural image dataset and a
  new dataset of complex questions about abstract shapes.
\end{abstract}

\section{Introduction} 

\begin{figure}[t] \begin{center}
    \includegraphics[width=\linewidth]{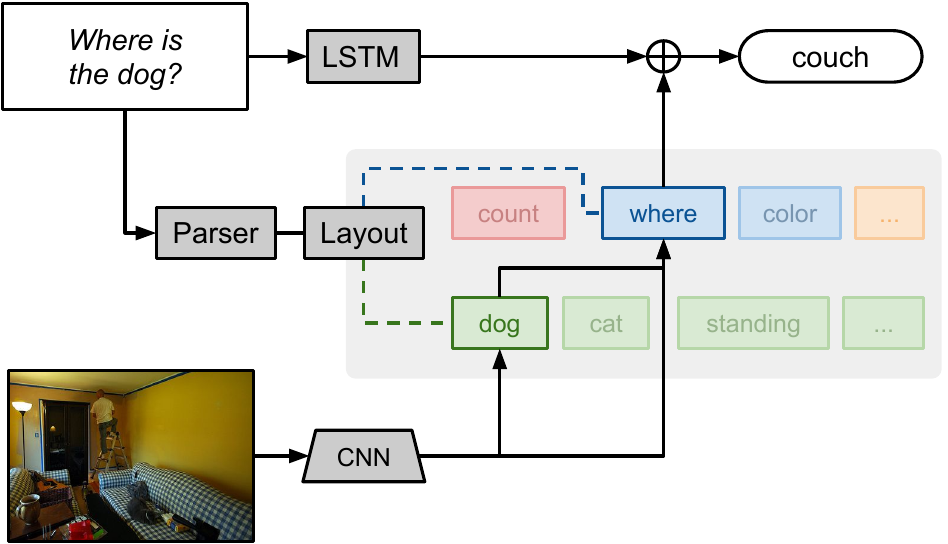} \end{center} \caption{
      A schematic representation of our proposed model---the shaded gray area is a
      {\it neural module network} of the kind introduced in this paper. Our
      approach uses a natural language parser to dynamically lay out a deep
      network composed of reusable modules. For visual question answering tasks,
      an additional sequence model provides sentence context and learns
      common-sense knowledge.
    } \label{fig:teaser}
\end{figure}

This paper describes an approach to visual question answering based on
\emph{neural module networks} (NMNs). We answer natural language questions about
images using collections of jointly-trained neural ``modules'', dynamically
composed into deep networks based on linguistic structure. 

Concretely, given an image and an associated question (e.g.\ \emph{where is
the dog?}), we wish to predict a corresponding answer (e.g.\ \emph{on the
couch}, or perhaps just \emph{couch}) (\autoref{fig:teaser}).
The visual QA task has significant significant applications to human-robot
interaction, search, and accessibility, and has been the subject of a great deal
of recent research attention
\cite{antol15iccv,gao2015you,ma15arxiv,malinowski15iccv,ren2015image,yu15arxiv}.
The task requires sophisticated understanding of both visual
scenes and natural language.
 Recent successful approaches represent
questions as bags of words, or encode the question using a recurrent neural
network \cite{malinowski15iccv} and train a simple classifier on the encoded
question and image. In contrast to these monolithic approaches, another line of
work for textual QA \cite{Liang13DCS} and image QA \cite{malinowski14nips} uses
semantic parsers to decompose questions into logical expressions. These logical
expressions are evaluated against a purely logical representation of the world,
which may be provided directly or extracted from an image
\cite{Krish2013Grounded}.

In this paper we draw from both lines of research, presenting a technique for
integrating the representational power of neural networks with the flexible
compositional structure afforded by symbolic approaches to semantics.  Rather
than relying on a monolithic network structure to answer all questions, our
approach assembles a network on the fly from a collection of specialized,
jointly-learned modules (\autoref{fig:teaser}). Rather than using logic to
reason over truth values, we remain entirely in the domain of visual features
and attentions.

Our approach first analyzes
each question with a semantic parser, and uses this analysis to
determine the basic computational units (attention, classification, etc.) needed
to answer the question, as well as the relationships between the modules. In
\autoref{fig:teaser}, we first produce an attention focused on the dog, which
passes its output to
a location classifier. Depending on the underlying structure, these messages
passed between modules may be raw image features, attentions, or classification
decisions; each module is determined by its input and output types.
Different kinds of modules are shown in different colors; attention modules
(like \mod{dog}) are shown in green, while labeling modules (like
\mod{where}) are shown in blue.
Importantly, all modules in an NMN are independent and composable, which allows
the computation to be different for each problem instance, and possibly
unobserved during training. 
Outside the NMN, our final answer 
uses a recurrent network (LSTM) to read the
question, which has been shown to be important to model common sense knowledge
and dataset biases \cite{malinowski15iccv}.

We evaluate our approach on two visual question answering tasks. On the
recently-released VQA \cite{antol15iccv} datasets we achieve results comparable
to or better than existing approaches, and show that our approach specifically
outperforms previous work on questions with compositional structure (\eg
requiring that an object be located and one of its attributes described). It
turns out, however, that many of the questions in both datasets are quite
simple, with little composition or reasoning required. To test our approach's
ability to handle harder questions, we introduce a new dataset of synthetic
images paired with complex questions involving spatial relations, set-theoretic
reasoning, and shape and attribute recognition.  On this dataset we outperform
competing approaches by as much as 25\% absolute accuracy.

While all the applications considered in this paper involve visual question
answering, the general architecture is potentially of broader usefulness, and
might be more generally applied to visual referring expression resolution
\cite{FitzGerald13Referring} or question answering about natural language texts
\cite{Iyyer14Factoid}.

To summarize our contributions: We first describe neural module networks, a
general architecture for discretely composing heterogeneous, jointly-trained
neural modules into deep networks. Next, for the visual QA task specifically, we
show how to construct NMNs based on the output of a semantic parser, and use
these to successfully complete established visual question answering tasks.
Finally, we introduce a new dataset of challenging, highly compositional
questions about abstract shapes, and show that our model again outperforms
previous approaches. We will release this dataset, as well as code for all
systems described in this paper, upon publication.

\section{Motivations}

We begin with two simple observations. First, state-of-the-art performance on
the full range of computer vision tasks that are studied requires a variety of
different deep network topologies---there is no single ``best network'' for all
tasks. Second, though different networks are used for different purposes, it is
commonplace to initialize systems for many of vision tasks with a prefix of a
network trained for classification \cite{Girshick14RCNN}.  This has been shown
to substantially reduce training time and improve accuracy. 

So while network structures are not \emph{universal} (in the sense that the same
network is appropriate for all problems), they are at least empirically
\emph{modular} (in the sense that intermediate representations for one task are
useful for many others). 

Can we generalize this idea in a way that is useful for question answering?
Rather than thinking of question answering as a problem of learning a single
function to map from questions and contexts to answers, it's perhaps useful to
think of it as a highly-multitask learning setting, where each problem instance
is associated with a novel task, and the identity of that task is expressed only
noisily in language. In particular, where a simple question like \emph{is this a
truck?} requires us to retrieve only one piece of information from an image,
more complicated questions, like \emph{how many objects are to the left of
the toaster?} might require multiple processing steps. The compositional nature
of language means that the number of such processing such steps is
potentially unbounded. Moreover, multiple \emph{kinds} of processing might be
required---repeated convolutions might identify a truck, but some kind of
recurrent architecture is likely necessary to count up to arbitrary numbers.

Thus our goal in this paper is to specify a framework for modular, composable,
jointly-trained neural networks. In this framework, we first predict the
structure of the computation needed to answer each question individually, then
realize this structure by constructing an appropriately-shaped neural network
from an inventory of reusable modules. These modules are learned jointly, rather
than trained in isolation, and specialization to individual tasks (identifying
properties, spatial relations, etc.) arises naturally from the training
objective.

\section{Related work}
We consider three lines of related work: previous efforts toward visual question
answering, discrete models for compositional semantics, and models that are
structurally similar to neural module networks.

\paragraph{Visual Question Answering}
Answering questions about images is sometimes referred to as a ``Visual
Turing Test'' \cite{malinowski14nips,geman15nas}. It has only recently gained
popularity, following the emergence of appropriate datasets consisting of paired
images, questions, and answers. While the DAQUAR dataset \cite{malinowski14nips}
is restricted to indoor scenes and contains relatively few examples, the \cocoqa
dataset \cite{yu15arxiv} and the VQA dataset \cite{antol15iccv} are
significantly larger and have more visual variety. Both are based on images from
the COCO dataset \cite{lin14eccv}. While \cocoqa contains question-answer pairs
automatically generated from the descriptions associated with the COCO dataset,
\cite{antol15iccv} has crowed sourced questions-answer pairs. We evaluate our
approach on VQA, the larger and more natural of the two datasets.

Notable ``classical'' approaches to this task include
\cite{malinowski14nips,Krish2013Grounded}. Both of these approaches are similar
to ours in their use of a semantic parser, but both rely on fixed logical
inference rather than learned compositional operations.

Several neural models for visual questioning have already been proposed in the
literature \cite{ren2015image,ma15arxiv,gao2015you}, all of which use standard
deep sequence modeling machinery to construct a joint embedding of image and
text, which is immediately mapped to a distribution over answers. Here we
attempt to more explicitly model the computational process needed to produce
each answer, but benefit from techniques for producing sequence and image
embeddings that have been important in previous work.

One important component of visual questioning is grounding the question in the
image. This grounding task has previously been approached in
\cite{karpathy14nips,plummer15iccv,karpathy15cvpr,kong14cvpr}, where the authors
tried to localize phrases in an image. \cite{xu2015arxiv} use an attention
mechanism, to predict a heatmap for each word, as an auxiliary task, during
sentence generation. The attentional component of our model is inspired by these approaches.

\paragraph{General compositional semantics}

There is a large literature on learning to answer questions about structured
knowledge representations from question--answer pairs, both with and without
joint learning of meanings for simple predicates \cite{Liang13DCS,Krish2013Grounded}.
Outside of question answering, several models have been proposed for instruction
following that impose a discrete ``planning structure'' over an underlying
continuous control signal \cite{Andreas14Paths,matuszek12icml}.  We are unaware
of past use of a semantic parser to predict network structures, or more
generally to exploit the natural similarity between set-theoretic approaches to
classical semantic parsing and attentional approaches to computer vision.

\paragraph{Neural network architectures}

The idea of selecting a different network graph for each input datum is
fundamental to both recurrent networks (where the network grows in the length of
the input) \cite{Elman90RNN} and recursive neural networks (where the network is
built, e.g., according to the syntactic structure of the input)
\cite{Socher13CVG}. But both of these approaches ultimately involve repeated
application of a single computational module (\eg an LSTM
\cite{Hochreiter97LSTM} or GRU \cite{Cho14GRU} cell).  From another direction,
some kinds of memory networks \cite{Weston14MemNet} may be viewed as a special
case of our model with a fixed computational graph, consisting of a sequence of
\mod{attend} modules followed by a \mod{classify} module (see
\autoref{sec:model} below).

Our basic contribution is in both assembling this graph on the fly, and
simultaneously in allowing the nodes to perform heterogeneous computations, with
for "messages" of different kinds---raw image features, attentions,
classification predictions---passed from one module to the next. We are unaware
of any previous work allowing such mixed collections of modules to be trained
jointly.

\section{Neural module networks for visual QA}
\label{sec:model}

Each training datum for this task can be thought of as a 3-tuple $(w, x, y)$, where
\begin{itemize}
  \setlength\itemsep{0em}
  \item $w$ is a natural-language question
  \item $x$ is an image
  \item $y$ is an answer
\end{itemize}
A model is fully specified by a collection of modules $\{ m \}$, each with
associated parameters $\theta_m$, and a \emph{network layout predictor} $P$
which maps from strings to networks. Given $(w, x)$ as above, the model
instantiates a network based on $P(w)$, passes $x$ (and possibly $w$ again) as
inputs, and obtains a distribution over labels (for the VQA task, we require the
output module to be a classifier). Thus a model ultimately encodes a predictive
distribution $p(y\ |\ w,\ x;\ \theta)$.

In the remainder of this section, we describe the set of modules used for the
VQA task, then explain the process by which questions are converted to network
layouts.

\subsection{Modules}

Our goal here is to identify a small set of modules that can be assembled into
all the configurations necessary for our tasks. This corresponds to identifying
a minimal set of composable vision primitives. The modules operate on three
basic data types: images, unnormalized attentions, and labels.  For the
particular task and modules described in this paper, almost all interesting
compositional phenomena occur in the space of attentions, and it is not
unreasonable to characterize our contribution more narrowly as an
``attention-composition'' network. Nevertheless, other types may be easily added
in the future (for new applications or for greater coverage in the VQA domain).

First, some notation: module names are typeset in a {\tt fixed width font}, and
are of the form \mod{TYPE[INSTANCE](ARG$_1$, \ldots)}. \mod{TYPE} is a
high-level module type (attention, classification, etc.) of the kind described
in this section. \mod{INSTANCE} is the particular instance of the model under
consideration---for example, \mod{attend[red]} locates red things, while
\mod{attend[dog]} locates dogs. Weights may be shared at both the type and
instance level. Modules with no arguments implicitly take the image as input;
higher-level arguments may also inspect the image.

\pagebreak
\subsubsection*{Attention}
\[
  \mod{attend} : \mathit{Image} \to \mathit{Attention}
\]
\includegraphics[width=\columnwidth]{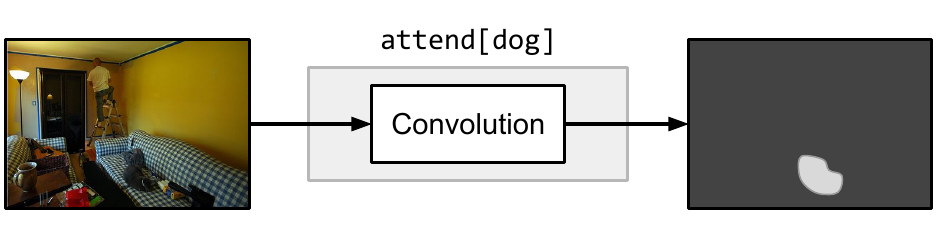}
An attention module \mod{attend[$c$]} convolves every position in the input
image with a weight vector (distinct for each $c$) to produce a heatmap or
unnormalized attention. So, for example, the output of the module {\small\tt
attend[dog]} is a matrix whose entries should be in regions of the image
containing cats, and small everywhere else, as shown above.\\

\subsubsection*{Re-attention}
\[
  \mod{re-attend} : \mathit{Attention} \to \mathit{Attention}
\]
\includegraphics[width=\columnwidth]{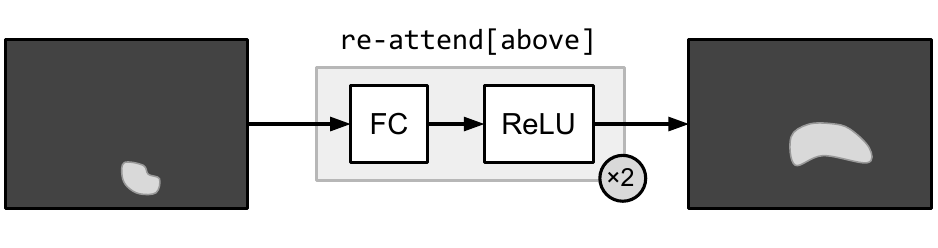}
A re-attention module \mod{re-attend[$c$]} is essentially just a multilayer
perceptron with rectified nonlinearities (ReLUs), performing a fully-connected
mapping from one attention to another.  Again, the weights for this mapping are
distinct for each $c$. So \mod{re-attend[above]} should take an attention and
shift the regions of greatest activation upward (as above), while
\mod{re-attend[not]} should move attention away from the active regions. For the
experiments in this paper, the first fully-connected (FC) layer produces a
vector of size 32, and the second is the same size as the input. \\

\subsubsection*{Combination}
\[
  \mod{combine} : \mathit{Attention} \times \mathit{Attention} \to
  \mathit{Attention}
\]
\includegraphics[width=\columnwidth]{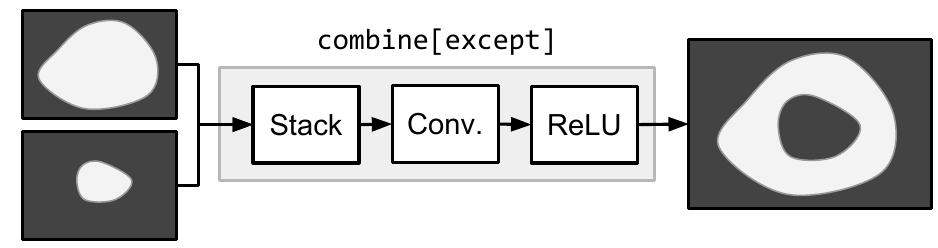}
A combination module \mod{combine[$c$]} merges two attentions into a single
attention. For example, {\small\tt combine[and]} should be active only in the
regions that are active in both inputs, while {\small\tt{combine[except]}}
should be active where the first input is active and the second is
inactive.

\subsubsection*{Classification}
\[
  \mod{classify} : \mathit{Image} \times \mathit{Attention} \to
  \mathit{Label}
\]
\includegraphics[width=\columnwidth]{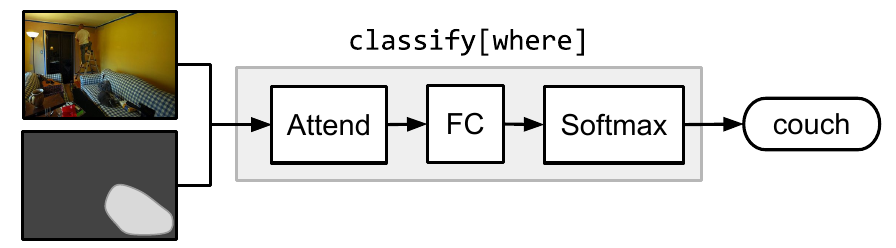}
A classification module \mod{classify[$c$]} takes an attention and the input
image and maps them to a distribution over labels. For example, {\small\tt
classify[color]} should return a distribution over colors in the region
attended to.

\subsubsection*{Measurement}
\[
  \mod{measure} : \mathit{Attention} \to \mathit{Label}
\]
\includegraphics[width=\columnwidth]{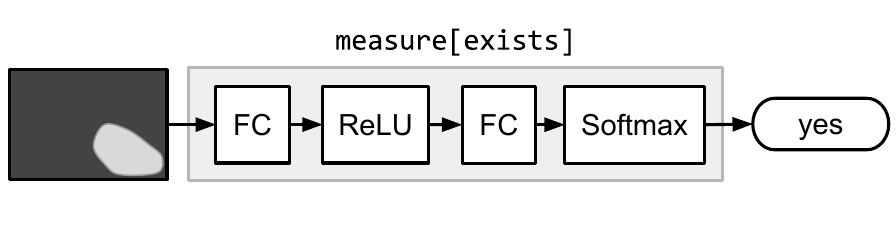}
A measurement module \mod{measure[$c$]} takes an attention alone and maps it to
a distribution over labels. Because attentions passed between modules are
unnormalized, \mod{measure} is suitable for evaluating the existence of a
detected object, or counting sets of objects.

\begin{figure*}
  \begin{subfigure}[t]{0.4\textwidth}
    \includegraphics[width=\textwidth]{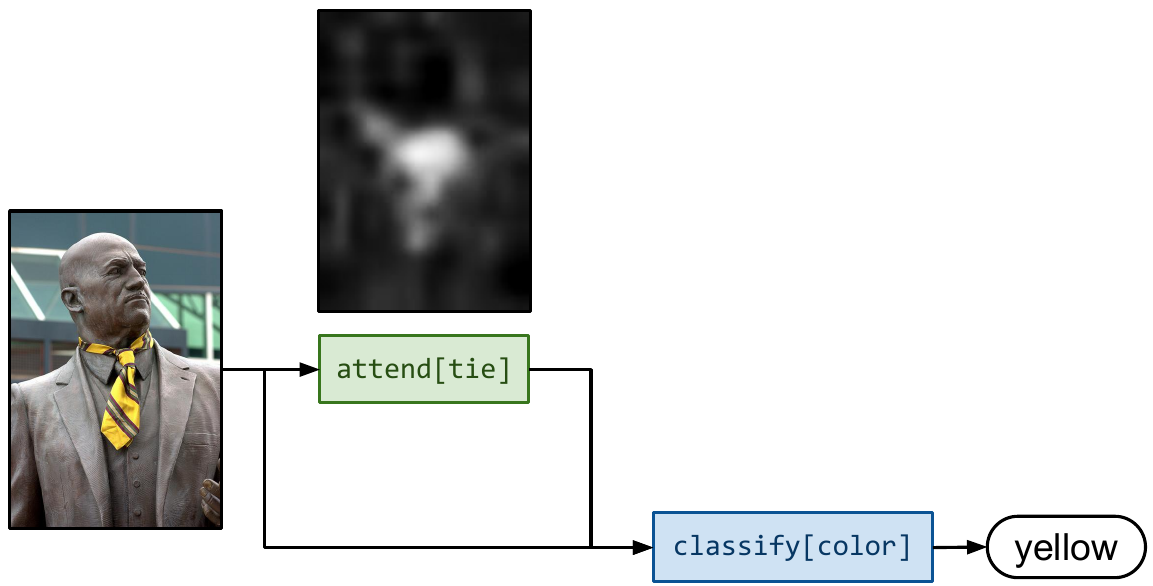}
    \caption{NMN for answering the question \emph{What color is
    his tie?} The \mod{attend[tie]} module first predicts a heatmap
    corresponding to the location of the tie. Next, the \mod{classify[color]}
    module uses this heatmap to produce a weighted average of image features,
  which are finally used to predict an output label.}
  \end{subfigure}
  \hfill
  \begin{subfigure}[t]{0.55\textwidth}
    \includegraphics[width=\textwidth]{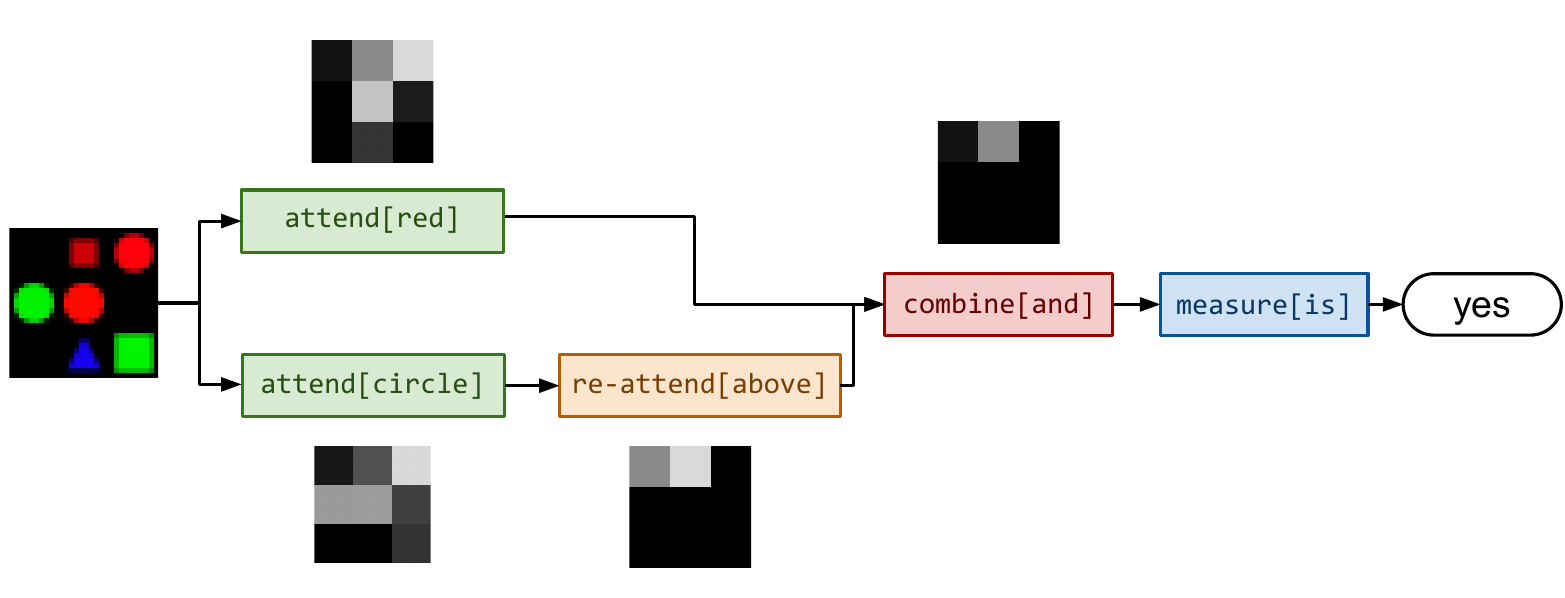}
    \caption{NMN for answering the question \emph{Is there a red shape above a
    circle?} The two \mod{attend} modules locate the red shapes and circles,
    the \mod{re-attend[above]} shifts the attention above the circles, the
    \emph{combine} module computes their intersection, and the
    \mod{measure[is]} module inspects the final attention and determines
  that it is non-empty.}
  \label{fig:shape-nmn}
  \end{subfigure}
  \caption{Sample NMNs for question answering about natural images and shapes.
  For both examples, layouts, attentions, and answers are real predictions made by
  our model.}
  \label{fig:nmns}
\end{figure*}

\subsection{From strings to networks}

Having built up an inventory of modules, we now need to assemble them into the
layout specified by the question.
The transformation from a natural language question to an instantiated neural
network takes place in two steps.  First we map from natural language questions
to \emph{layouts}, which specify both the set of modules used to answer a given
question, and the connections between them. Next we use these layouts are used
to assemble the final prediction networks.

We use standard tools pre-trained on existing linguistic resources to obtained
structured representations of questions. Future work might focus on learning (or
at least fine-tuning) this prediction process jointly with the rest of the
system. %

\paragraph{Parsing}
We begin by parsing each question with the Stanford Parser
\cite{Klein03Unlex}.
to obtain a universal dependency representation \cite{DeMarneffe08Deps}.
Dependency parses express grammatical relations between parts of a sentence
(e.g.\ between objects and their attributes, or events and their participants),
and provide a lightweight abstraction away from the surface form of the
sentence. The parser also performs basic lemmatization, for example turning
\emph{kites} into \emph{kite} and \emph{were} into \emph{be}. This reduces
sparsity of module instances.

Next, we filter the set of dependencies to those connected the wh-word in the
question (the exact distance we traverse varies depending on the task). This
gives a simple symbolic form expressing (the primary) part of the sentence's
meaning.  For example, \emph{what is standing in the field} becomes
\mod{what(stand)}; \emph{what color is the truck} becomes \mod{color(truck)},
and \emph{is there a circle next to a square} becomes \mod{is(circle,
next-to(square))}.
In the process we also strip away function words like determiners and
modals, so \emph{what type of cakes were they?} and \emph{what type of cake is
it?} both get converted to \mod{type(cake)}.
The code for transforming parse trees to structured queries
will be provided in the accompanying software package.

These representations bear a certain resemblance to pieces of a combinatory
logic \cite{Liang13DCS}: every leaf is implicitly a function taking the image as
input, and the root represents the final value of the computation.  But our
approach, while compositional and combinatorial, is crucially not logical: the
inferential computations operate on continuous representations produced by
neural networks, becoming discrete only in the prediction of the final answer.

\paragraph{Layout}
These symbolic representations already determine the structure of the predicted
networks, but not the identities of the modules that compose them. This final
assignment of modules is fully determined by the structure of the parse. All
leaves become \mod{attend} modules, all internal nodes become \mod{re-attend} or
\mod{combine} modules dependent on their arity, and root nodes become
\mod{measure} modules for yes/no questions and \mod{classify} modules for all
other question types.

Given the mapping from queries to network layouts described above, we have for
each training example a network structure, an input image, and an output label.
In many cases, these network structures are different, but have tied parameters.
Networks which have the same high-level structure but different instantiations
of individual modules (for example \emph{what color is the cat?}---{\small\tt
classify[color](attend[cat])} and \emph{where is the truck?}---{\small\tt
classify[where](attend[truck])}) can be processed in the same batch, resulting
in efficient computation.

As noted above, parts of this conversion process are task-specific---we found
that relatively simple expressions were best for the natural image questions,
while the shapes question (by design) required deeper structures. Some summary
statistics are provided in \autoref{tab:struct-summary}.

\begin{table*}
  \centering
  \begin{tabular}{cccccc}
    \toprule
    & types & \# instances & \# layouts & max depth & max size \\
    \midrule
    VQA & \footnotesize\tt attend, combine, classify, measure & 1995 & 66549 & 3 & 4 \\
    \shapes & \footnotesize\tt attend, re-attend, combine, measure & 8 & 164 & 5 & 6 \\
    \bottomrule
  \end{tabular}
  \caption{Structure summary statistics for neural module networks used in this
    paper. ``types'' is the set of high-level module types available (e.g.
    \texttt{attend}), ``\# instances'' is the number of specific module instances
    (e.g. \mod{attend[llama]}), and ``\# layouts'' is the number of distinct
    composed structures (e.g. \mod{classify[color](attend[llama])}).
    ``Max depth'' is the greatest depth across all layouts, while ``max
    size'' is the greatest number of modules---for example, the network in
    \autoref{fig:shape-nmn}
    has depth 4 and size 5. (All numbers from training sets.)
  }
  \label{tab:struct-summary}
  \vspace{-.5em}
\end{table*}

\paragraph{Generalizations}

It is easy to imagine applications where the input to the layout stage comes
from something other than a natural language parser. Users of an image database,
for example, might write SQL-like queries directly in order to specify their
requirements precisely, e.g.
\begin{flushleft}
  {\tt IS(cat) AND NOT(IS(dog))}
\end{flushleft}
or even mix visual and non-visual specifications in their queries:
\begin{flushleft}
  {\tt IS(cat) and date > 2014-11-5}
\end{flushleft}

Indeed, it is possible to construct this kind of ``visual SQL'' using precisely
the approach described in this paper---once our system is trained, the learned
modules for attention, classification, etc. can be assembled by any kind of
outside user, without relying on natural language specifically.

\subsection{Answering natural language questions}

So far our discussion has focused on the neural module net architecture, without
reference to the remainder of \autoref{fig:teaser}. Our final model combines the output
from the neural module network with predictions from a simple LSTM question
encoder. This is important for two reasons. First, because of the relatively
aggressive simplification of the question that takes place in the parser,
grammatical cues that do not substantively change the semantics of the question,
but which might affect the answer, are discarded. For example, \emph{what is
flying?} and \emph{what are flying?} both get converted to \mod{what(fly)}, but
their answers should be \emph{kite} and \emph{kites} respectively, even given
the same underlying image features. The question encoder thus allows us to model
underlying \emph{syntactic} regularities in the data. Second, it allows us to
capture \emph{semantic} regularities: with missing or low-quality image data, it
is reasonable to guess that \emph{what color is the bear?} is answered by
\emph{brown}, and unreasonable to guess \emph{green}. The question encoder also
allows us to model effects of this kind.

All experiments in this paper use a standard single-layer LSTM with 1024 hidden
units. The question modeling component predicts a distribution over the set of
answers, like the root module of the NMN. The final prediction from the model is
a geometric average of these two probability distributions, dynamically
reweighted using both text and image features. The complete model, including
both the NMN and sequence modeling component, is trained jointly.

\section{Training neural module networks}

Our training objective is simply to find module parameters maximizing the
likelihood of the data. By design, the last module in every network outputs a
distribution over labels, and so each assembled network also represents a
probability distribution.

Because of the dynamic network structures used to answer questions, some weights
are updated much more frequently than others. For this reason we found that
learning algorithms with adaptive per-weight learning rates performed
substantially better than simple gradient descent. All the experiments described
below use AdaDelta \cite{Zeiler12Adadelta}  (thus there was no hyperparameter
search over step sizes).

It is important to emphasize that the labels we have assigned to distinguish
instances of the same module type---\mod{cat}, \mod{and}, etc.---are a
notational convenience, and do not reflect any manual specification of the
behavior of the corresponding modules. \mod{detect[cat]} is not fixed or even
initialized %
as cat recognizer (rather than a couch recognizer or a dog recognizer), and
\mod{combine[and]} isn't fixed to compute intersections of attentions (rather
than unions or differences). Instead, they acquire these behaviors as a
byproduct of the end-to-end training procedure. As can be seen in
\autoref{fig:nmns},
the image--answer pairs and parameter tying together encourage each module to
specialize in the appropriate way.

\section{Experiments: compositionality}

We begin with a set of motivating experiments on synthetic data.
Compositionality, and the corresponding ability to answer questions with
arbitrarily complex structure, is an essential part of the kind of deep image
understanding visual QA datasets are intended to test. At the same time,
questions in most existing natural image datasets are quite simple, for the most
part requiring that only one or two pieces of information be extracted from an
image in order to answer it successfully, and with little evaluation of
robustness in the presence of distractors (e.g. asking \emph{is there a blue
house} in an image of a red house and a blue car).

As one of the primary goals of this work is to learn models for deep semantic
compositionality, we have created \shapes, a synthetic dataset that places such
compositional phenomena at the forefront. This dataset consists of complex
questions about simple arrangements of colored shapes (\autoref{fig:examples}).  Questions contain
between two and four attributes, object types, or relationships.  There are 244
questions and 15616 images in total. %
To eliminate mode-guessing as a
viable strategy, all questions have a yes-or-no answer, but good performance
requires that the system learn to recognize shapes and colors, and understand
both spatial and logical relations among sets of objects.

While success on this dataset is by no means a sufficient condition for robust
visual QA, we believe it is a necessary one. In this respect it is similar in
spirit to the bAbI \cite{Weston15BABI} dataset, and we hope that \shapes will
continue to be used in conjunction with natural image datasets.

To produce an initial set of image features, we pass the input image through the
convolutional portion of a LeNet \cite{LeCun90LeNet} which is jointly trained
with the question-answering part of the model. We compare our approach to a
reimplementation of the VIS+LSTM baseline similar to the one described by
\cite{Ren15VQA}, again swapping out the pre-trained image embedding with a
LeNet.

\begin{table}
  \footnotesize
  \centering
  \begin{tabular}{lcccc}
    \toprule
    & size 4 & size 5 & size 6 & All \\
    \midrule
    Majority & 64.4 & 62.5 & 61.7 & 63.0 \\
    VIS+LSTM & 71.9 & 62.5 & 61.7 & 65.3 \\
    NMN      & 89.7 & 92.4 & 85.2 & \bf 90.6 \\
    NMN (easy) & 97.7 & 91.1 & 89.7 & \bf 90.8 \\
    \bottomrule
  \end{tabular}
  \caption{Results on the \shapes dataset. Here ``size'' is the number of
    modules needed to instantiate an appropriate NMN. Our model achieves high
    accuracy and outperforms a baseline from previous work, especially on highly
    compositional questions. ``NMN (easy)''  is a modified training set with no
    size-6 questions; these results demonstrate that our model is able to
    generalize to questions more complicated than it has ever seen at training
    time.
  }
  \label{tab:shapes-results}
  \vspace{-.5em}
\end{table}

As can be seen in \autoref{tab:shapes-results}, our model achieves excellent
performance on this dataset, while the VIS+LSTM baseline fares little better
than a majority guesser. Moreover, the color detectors and attention
transformations behave as expected (\autoref{fig:shape-nmn}), indicating that
our joint training procedure correctly allocates responsibilities among modules.
This confirms that our approach is able to model complex compositional phenomena
outside the capacity of previous approaches to visual question answering. 

We perform an additional experiment on a modified version of the training set,
which contains no size-6 questions (i.e. questions whose corresponding NMN has 6
modules). Here our performance does not suffer at all, and perhaps increases
slightly; this demonstrates that our model is able to generalize to questions
even more complicated than those it has seen during training. Using only
linguistic information, the model extrapolates simple visual patterns it has
learned to even harder questions.

\section{Experiments: natural images}

\begin{figure*}
  \vspace{-1em}
  \renewcommand{\arraystretch}{1.5}
  \centering
  \footnotesize
  \begin{tabular}{|>{\raggedright}*{5}{p{.17\textwidth}|}}
    \hline
    & & & & \\[-.7em]
    \includegraphics[width=\linewidth]{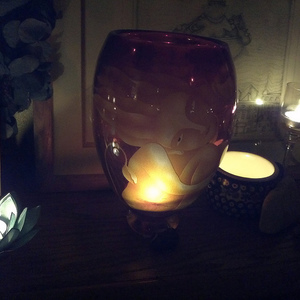} &
    \includegraphics[width=\linewidth]{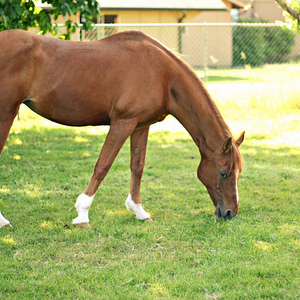} &
    \includegraphics[width=\linewidth]{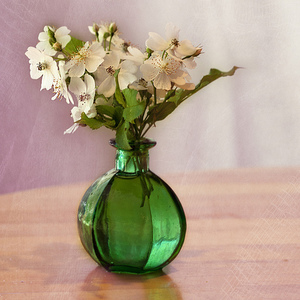} &
    \includegraphics[width=\linewidth]{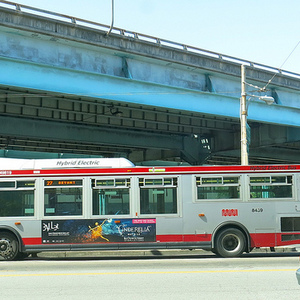} &
    \includegraphics[width=\linewidth]{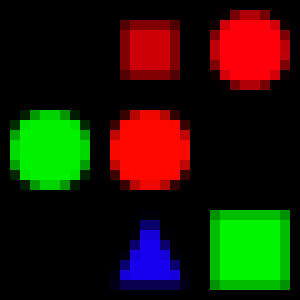} \\
    \emph{how many different lights in various different shapes and sizes?} &
    \emph{what is the color of the horse?} &
    \emph{what color is the vase?} &
    \emph{is the bus full of passengers?} &
    \emph{is there a red shape above a circle?} \\
    & & & & \\[-.7em]
    \hline
    \\[-3em]
\begin{verbatim}measure[count](
  attend[light])\end{verbatim} &
\begin{verbatim}classify[color](
  attend[horse])\end{verbatim} &
\begin{verbatim}classify[color](
  attend[vase])\end{verbatim} &
\begin{verbatim}measure[is](
  combine[and](
    attend[bus],
    attend[full])\end{verbatim} &
\begin{verbatim}measure[is](
  combine[and](
   attend[red],
   re-attend[above](
    attend[circle])))\end{verbatim}
    \\[-.5em]
    \hline
    four (four) &
    brown (brown) &
    green (green) &
    yes (yes) &
    no (no) \\
    \hline
  \end{tabular}\\[1.3em]
  \begin{tabular}{|>{\raggedright}*{5}{p{.17\textwidth}|}}
    \hline
    & & & & \\[-.7em]
    \includegraphics[width=\linewidth]{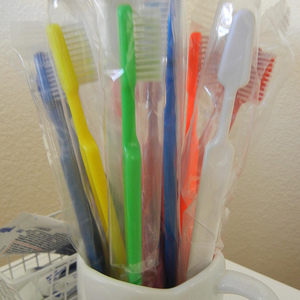} &
    \includegraphics[width=\linewidth]{fig/thumb/cocoqa4.jpg} &
    \includegraphics[width=\linewidth]{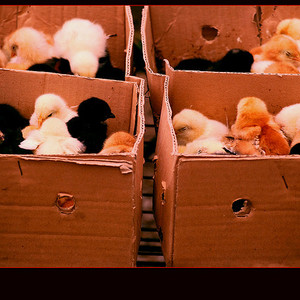} &
    \includegraphics[width=\linewidth]{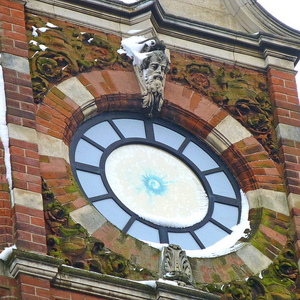} &
    \includegraphics[width=\linewidth]{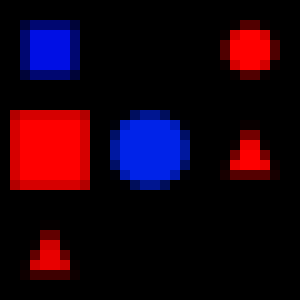} \\
    \emph{what is stuffed with toothbrushes wrapped in plastic?} & 
    \emph{where does the tabby cat watch a horse eating hay?} &
    \emph{what material are the boxes made of?} &
    \emph{is this a clock?} &
    \emph{is a red shape blue?} \\
    & & & & \\[-.7em]
    \hline
    \\[-3em]
\begin{verbatim}classify[what](
  attend[stuff])\end{verbatim} &
\begin{verbatim}classify[where](
  attend[watch])\end{verbatim} &
\begin{verbatim}classify[material](
  attend[box])\end{verbatim} &
\begin{verbatim}measure[is](
attend[clock])\end{verbatim} &
\begin{verbatim}measure[is](
  combine[and](
    attend[red],
    attend[blue]))\end{verbatim}
\\[-.5em]
 \hline
    container (cup) &
    pen (barn) &
    leather (cardboard) &
    yes (no) &
    yes (no) \\
    \hline
  \end{tabular}
  \caption{Example output from our approach on different visual QA tasks. The
  top row shows correct answers, while the bottom row shows mistakes (correct
  answers are given in parentheses).}
  \label{fig:examples}
\end{figure*}

Next we consider the model's ability to handle hard perceptual problems
involving natural images.  Here we evaluate on the recently-released VQA
dataset. This is the largest resource of its kind, consisting of more than
200,000 images, each paired with three questions and ten answers per question.
Data was generated by human annotators, in contrast to previous work, which has
generated questions automatically from captions \cite{Ren15VQA}.  We learn our
model using the standard train/test split, training only with those answers
marked as high confidence.  The visual input to the NMN is the conv5 layer of a
16-layer VGGNet \cite{Simonyan14VGG} after max-pooling. We do not fine-tune the
VGGNet.

Results are shown in \autoref{tab:vqa-results}. As can be seen, we outperform
the best published results on this task. A breakdown of our questions by answer
type reveals that our model performs especially well on questions answered by an
object, attribute, or number, but worse than a sequence baseline in the yes/no
category. Inspection of training-set accuracies suggests that performance on
yes/no questions is due to overfitting. 
An ensemble with a sequence-only system might achieve even better results;
future work within the NMN framework should focus on redesigning the
\mod{measure} module to reduce effects from overfitting.

Inspection of parser outputs also suggests that there is substantial room to
improve the system using a better parser. A hand inspection of the first 50
parses in the training set suggests that most (80--90\%) of questions asking for
simple properties of objects are correctly analyzed, but more complicated
questions are more prone to picking up irrelevant predicates. For example
\emph{are these people most likely experiencing a work day?} is parsed as
\mod{be(people, likely)}, when the desired analysis is \mod{is(people, work)}.
Parser errors of this kind could be fixed with joint learning.

\autoref{fig:examples} is broadly suggestive of the kinds of prediction errors
made by the system, including plausible semantic confusions (cardboard
interpreted as leather, round windows interpreted as clocks), normal lexical
variation (\emph{container} for \emph{cup}), and use of answers that are \emph{a
priori} plausible but unrelated to the image (describing a horse as located in a
pen rather than a barn).

\begin{table}
  \footnotesize
  \center
  \begin{tabular}{lccccc}
    \toprule
    & \multicolumn{4}{c}{test-dev} & test \\
    \cmidrule(lr){2-5} \cmidrule(lr){6-6}
    & Yes/No & Number & Other & All & All \\
    \midrule
    LSTM \cite{antol15iccv} & 78.20 & 35.7 & 26.6 & 48.8 & -- \\
    VIS+LSTM \cite{antol15iccv} & 78.9 & 35.2 & 36.4 & 53.7 & 54.1 \\
    NMN & 69.38 & 30.7 & 22.7 & 42.7 & -- \\
    NMN+LSTM & 77.7 & 37.2 & 39.3 & 54.8 & \bf 55.1 \\
    \bottomrule
  \end{tabular}
  \caption{Results on the VQA test server. NMN+LSTM is the full model shown in
    \autoref{fig:teaser}, while NMN is an ablation experiment with no
    whole-question LSTM. The full model outperforms previous
    approaches, scoring particularly well on questions not involving a
    binary decision. Baseline numbers are as reported in previous work.}
    \label{tab:vqa-results}
    \vspace{-1em}
\end{table}

\section{Conclusions and future work}

In this paper, we have introduced \emph{neural module networks}, which provide a
general-purpose framework for learning collections of neural modules which can
be dynamically assembled into arbitrary deep networks. We have demonstrated that
this approach achieves state-of-the-art performance on existing datasets for
visual question answering, performing especially well on questions answered by
an object or an attribute. Additionally, we have introduced a new dataset of
highly compositional questions about simple arrangements of shapes, and shown
that our approach substantially outperforms previous work.

So far we have maintained a strict separation between predicting network
structures and learning network parameters. It is easy to imagine that these two
problems might be solved jointly, with uncertainty maintained over network
structures throughout training and decoding. This might be accomplished either
with a monolithic network, by using some higher-level mechanism to ``attend'' to
relevant portions of the computation, or else by integrating with existing tools
for learning semantic parsers \cite{Krish2013Grounded}.

The fact that our neural module networks can be trained to produce predictable
outputs---even when freely composed---points toward a more general paradigm of
``programs'' built from neural networks. In this paradigm, network designers
(human or automated) have access to a standard kit of neural parts from which to
construct models for performing complex reasoning tasks. While visual question
answering provides a natural testbed for this approach, its usefulness is
potentially much broader, extending to queries about documents and structured
knowledge bases or more general signal processing and function approximation.

\section*{Acknowledgments}

The authors are grateful to Lisa Anne Hendricks, Eric Tzeng, and Russell Stewart
for useful conversations, and to Nvidia for a hardware grant. JA is supported by
a National Science Foundation Graduate Research Fellowship.  MR is supported by
a fellowship within the FIT weltweit-Program of the German Academic Exchange
Service (DAAD).  This work was additionally supported by DARPA, AFRL, DoD MURI
award N000141110688, NSF awards IIS-1427425 and IIS-1212798, and the Berkeley
Vision and Learning Center.

\small
\bibliographystyle{ieee}
\bibliography{biblioShort,rohrbach,related,jacob}

\end{document}